\documentclass{article}

\usepackage[final]{neurips_2022}

\usepackage[utf8]{inputenc} 
\usepackage[T1]{fontenc}    
\usepackage{hyperref}       
\usepackage{url}            
\usepackage{booktabs}       
\usepackage{amsfonts}       
\usepackage{nicefrac}       
\usepackage{microtype}      
\usepackage{xcolor}         

\usepackage{amsmath}
\usepackage{amssymb}
\usepackage{graphicx}

\usepackage[capitalize,noabbrev]{cleveref}

\newcommand{\vb}{{\boldsymbol{b}}}

\newcommand{\ve}{{\boldsymbol{e}}}

\newcommand{\vi}{{\boldsymbol{i}}}

\newcommand{\vm}{{\boldsymbol{m}}}

\newcommand{\vo}{{\boldsymbol{o}}}

\newcommand{\vr}{{\boldsymbol{r}}}
\newcommand{\vs}{{\boldsymbol{s}}}

\newcommand{\vv}{{\boldsymbol{v}}}

\newcommand{\vx}{{\boldsymbol{x}}}
\newcommand{\vy}{{\boldsymbol{y}}}
\newcommand{\vz}{{\boldsymbol{z}}}

\newcommand{\vA}{{\boldsymbol{A}}}

\newcommand{\vU}{{\boldsymbol{U}}}

\newcommand{\vW}{{\boldsymbol{W}}}

\newcommand{\vzeta}{{\boldsymbol{\zeta}}}

\newcommand{\vtheta}{{\boldsymbol{\theta}}}

\newcommand{\vTheta}{{\boldsymbol{\Theta}}}

\newcommand{\trans}{^\mathsf{T}}

\newcommand{\fig}[1]{Fig.~\ref{#1}}
\newcommand{\eq}[1]{Eq.~\ref{#1}}

\title{Learning Dynamics and Structure of Complex Systems Using Graph Neural Networks}

\author{
	Zhe Li \textsuperscript{$\dagger$,1,3} \thanks{\textsuperscript{1} Department of Neuroscience, Baylor College of Medicine. \textsuperscript{2} Department of Electrical and Computer Engineering, Rice University. \textsuperscript{3} Center for Neuroscience and Artificial Intelligence, Baylor College of Medicine} \\
	\texttt{zhel@bcm.edu}
	\And
	Andreas S. Tolias \textsuperscript{1,2,3}
	\And
	Xaq Pitkow \textsuperscript{$\dagger$,1,2,3} \\
	\texttt{xaq@rice.edu}
}

\begin{document}

\maketitle

\begin{abstract}
	Many complex systems are composed of interacting parts, and the underlying laws are usually simple and universal. While graph neural networks provide a useful relational inductive bias for modeling such systems, generalization to new system instances of the same type is less studied. In this work we trained graph neural networks to fit time series from an example nonlinear dynamical system, the belief propagation algorithm. We found simple interpretations of the learned representation and model components, and they are consistent with core properties of the probabilistic inference algorithm. We successfully identified a `graph translator' between the statistical attributes in belief propagation and parameters of the corresponding trained network, and showed that it enables two types of novel generalization: to recover the underlying structure of a new system instance based solely on time series observations, and to construct a new network from this structure directly. Our results demonstrated a path towards understanding both dynamics and structure of a complex system and how such understanding can be used for generalization.
\end{abstract}

\section{Introduction}

Many real world problems involve dynamical systems composed of interacting parts, such as planet movement, social networks, protein folding, neural circuits, electrical grids, etc. While a system can show complex and rich behavior, the elementary rules about the interaction are usually much simpler. The ability to abstract these simple rules from complex dynamics is one hallmark of intelligence, as it is a key feature of understanding.

When modeling such a dynamical system, we can usually exploit the underlying symmetry and impose certain canonical assumptions on the model. Graph neural networks (GNN) explicitly model each part of the system, and use local messages to model the interaction among them \citep{Scarselli2009,Li2016,Battaglia2018,Ying2018}. While GNN framework has been used to study a variety of dynamical systems \citep{Gilmer2017,Watters2017,Kipf2018}, commonly it is used for modeling one specific instance of the system \citep{Battaglia2016,Bapst2020}. As a result it is not trivial to disentangle the different aspects of the learned system, and interpret components of the trained model \citep{Cranmer2020}. In this study, we propose to model multiple instances of the same type of system, and explicitly learn the universal dynamics which are shared across all instances, as well as the specific structure of each individual one. We analyze to disentangle these two aspects of the learned GNN models and show how this understanding enables generalization to new problem instances.

We choose the belief propagation (BP) algorithm on probabilistic graphical model (PGM) as the dynamical system of interest. BP performs probabilistic inference on a set of random variables and their statistical interactions via iterative computations. The algorithm itself is an important operation as it marginalizes out nuisance variables efficiently (approximately for loopy graphs), which is useful in many situations. Given the time series of BP outputs, we aim to understand how they change over time (dynamics) and to recover the underlying PGM (structure).

In this report, we begin by introducing the mathematical definition of GNNs used in our work. We next show the results of fitting GNN models to the BP outputs on time-varying multivariate Gaussian distributions, and analyze the representation and components of trained models. We then develop `graph translator' that links between the structural parameters of GNN models and static properties of Gaussian distributions, and show how it enables two types of generalization to new PGM instances. Lastly, we discuss the limitations and future directions of our approach.

\section{Background}

\subsection{Graph neural network}

A graph neural network \citep{Scarselli2009,Li2016,Battaglia2018} is a message-passing algorithm defined over a graph $\mathcal{G}=(\mathcal{V}, \mathcal{E})$ with vertices $\mathcal{V}$ and edges $\mathcal{E}$. For a graph of size $N$, $\mathcal{V}=\{1, \ldots, N\}$ and $\mathcal{E} \subseteq \{(i, j) \mid {(i, j)\in \mathcal{V}^2 \wedge i \neq j}\}$ where $(i, j)$ represents a directed edge from vertex $j$ to $i$. Each vertex $i$ is associated with a vertex parameter $\vv_i$, and each edge $(i, j)$ is associated with an edge parameter $\ve_{ij}$. We assume that all vertex parameters $\vv_i$ are of the same dimension $D_v$, and all edge parameters $\ve_{ij}$ are of dimension $D_e$.

We use GNNs to model observations of a dynamical system at discrete time points. At each time point $t$, vertex $i$ receives a $D_x$-dimensional external input $\vx_i^t$. We take $D_x$ to be the same for all vertices, and $D_x=0$ represents purely autonomous dynamics of a system. The state of vertex $i$ at time $t$ is denoted as $\vs_i^t$, which is a vector of dimension $D_s$. Vertex states get updated by both time-varying inputs (when $D_x>0$) and the lateral interaction from neighboring vertices in the form of messages.

Along the edge $(i, j)$, a pairwise message $\vm_{ij}^t$ of dimension $D_m$ is generated by a message function $\mathcal{M}(\cdot)$,
\begin{equation}
	\vm_{ij}^t = \mathcal{M}(\vs_{i}^t, \vs_{j}^t; \ve_{ij}, \vTheta_\mathcal{M}), \label{eq:msg.func}
\end{equation}
where $\vTheta_\mathcal{M}$ is the parameter of $\mathcal{M}(\cdot)$. All incoming messages on vertex $i$ are then aggregated via an aggregation function $\mathcal{A}(\cdot)$,
\begin{equation}
	\vm_i^t = \mathcal{A}(\{\vm_{ij}^t \mid (i, j) \in \mathcal{E}\}). \label{eq:aggr.func}
\end{equation}
$\mathcal{A}(\cdot)$ takes the set of pairwise messages as input, ignoring their order, hence it is invariant to permutation. Simple choices of $\mathcal{A}(\cdot)$ include element-wise summation and max pooling, while more complicated mechanisms such as attention-based aggregation can also be used \citep{Velickovic2018,liao2019efficient}. We choose summation in this study.
The vertex state is updated by an update function $\mathcal{U}(\cdot)$,
\begin{equation}
	\vs_i^{t+1} = \mathcal{U}(\vs_i^{t}, \vx_i^t, \vm_i^t; \vv_i, \vTheta_\mathcal{U}), \label{eq:upd.func}
\end{equation}
where $\vTheta_\mathcal{U}$ is the parameter of $\mathcal{U}(\cdot)$. And finally, the GNN output is the projection of states via a readout function $\mathcal{R}(\cdot)$ that applies to every vertex separately,
\begin{equation}
	\vo_i^t = \mathcal{R}(\vs_i^t; \vTheta_\mathcal{R}),
	\label{eq:canon.read}
\end{equation}
where $\vTheta_\mathcal{R}$ is the parameter of $\mathcal{R}(\cdot)$.

There is an intrinsic degeneracy among $\mathcal{M}(\cdot)$, $\mathcal{U}(\cdot)$ and $\mathcal{R}(\cdot)$: the complexity of one function may be compensated by the other two. In this study, we use simple linear readouts, so all nonlinearities must be captured by message and update functions. We also assume the readout function $\mathcal{R}(\cdot)$ does not depend on vertex parameters $\vv_i$.

\subsection{Belief propagation on probabilistic graph models}

A probabilistic graph model (PGM) describes a joint distribution of $N$ random variables $\theta_1, \ldots, \theta_N$. Here we focus on PGMs with pairwise interactions. In a graph of size $N$, each vertex $i$ is associated with a singleton potential $\phi_i(\theta_i)$, and each undirected edge $(i, j)$ is associated with a pairwise potential $\psi_{ij}(\theta_i, \theta_j)$. The joint distribution of $\left(\theta_1, \ldots, \theta_N\right)$ is proportional to the product of all potentials,
\begin{equation}
	p(\theta_1, \ldots, \theta_N) \propto \prod_{i\in\mathcal{V}} \phi_i(\theta_i) \cdot \prod_{(i, j)\in\mathcal{E}} \psi_{ij}(\theta_i, \theta_j).
\end{equation}

Marginalization of the joint distribution is one of the most used computation in statistical analysis, and many different methods have been developed for it, including belief propagation (BP), which is an iterative algorithm that operates on a PGM. At each time point $t$, BP estimates the marginal distribution $p_i(\theta_i)$ as $\hat{p}_i^t(\theta_i)$. BP is not guaranteed to converge, and even when it converges it is not guaranteed to converge to the true marginal distribution, however it provides a decent estimation in many cases \citep{Murphy1999,Wainwright2003}.

To make the BP dynamics more interesting, we use time-varying singleton potentials $\phi_i^t(\theta_i)$ while keeping the pairwise potential $\psi_{ij}(\theta_i, \theta_j)$ constant. This simulates a system with time-varying local evidence while the coupling among different parts is fixed. BP output $\hat{p}_i^t(x_i)$ therefore is the continuous estimate of $p_i^t(\theta_i)$ determined by $\phi_i^t(\theta_i)$ and $\psi_{ij}(\theta_i, \theta_j)$.

\section{Related work}

GNNs have been widely used to model time series of physical systems, often about predicting trajectories of interacting objects. In \citet{Chang2017,Bapst2020,SanchezGonzalez2020}, the adjacency matrix of the interaction graph is determined by spatial proximity instead of being learned from data, therefore are not applicable for systems in which `neighborhood' is not clearly defined (for instance probabilistic inference). \citet{Chang2017} demonstrates GNN can extrapolate to new environments containing more objects, but the structural properties of the system, namely the mass, is the same for all objects and environments.

\citet{Chang2017} uses GNNs to infer latent properties such as object mass. Similarly, \citet{Kipf2018} uses GNN to infer interaction type in a multi-object system. However, the vertex and edge properties in these studies are designed as discrete values. Our work instead investigates a spectrum of continuous graph properties and reveals the intrinsic low-dimensional structure of them.

\citet{Cranmer2020} encourages a GNN model to learn compact internal representation, and performs symbolic regression based on it. Our work reveals low-dimensional structure of GNNs also through regularization. We analyze states and messages as vectors instead of just analyzing individual components, hence avoid the loss of information when representation is not perfectly factorized. Also, we manage to disentangle the structural properties of the system (\textit{e.g.} mass, charge) instead of assuming they are known. The graph translator proposed in our work can be seen as a more general form of symbolic regression which enables generalization to new instances of complex system.

Besides modeling physical systems, past work has used GNNs for probabilistic inference \citep{pmlr-v97-qu19a,garcia2019combining}. GNN is often proposed as a better alternative to traditional BP algorithm, especially on loopy graphs \citep{yoon2019inference} or when higher-order statistics is critical \citep{zhang2019factor,fei2021generalization}. Our work instead focuses on using GNN to model BP algorithm as a dynamical system, whether it produces accurate probabilistic inference is outside the scope of this work.

\section{Methods}

\subsection{Multi-graph training} \label{sec:multi.train}

An important merit of GNN is that the dynamical and structural aspects of a system are represented separately. The canonical functions characterize the dynamics ``law'' of a certain type of system, such as the BP algorithm, while the graph parameters describe the structure of a particular system instance, \textit{e.g.} the pairwise coupling in a PGM.

Because both aspects affect the system behavior, it is not trivial to disentangle them based on the observations of the system. In order to inspect the effect of different structures while keeping the dynamics fixed, we need a large number of vertices and edges that cover a wide range of parameters, which suggests using a huge graph. However, training GNN on a huge fully connected graph is computationally expensive. Instead, a more feasible approach is to train multiple medium graphs simultaneously. It is equivalent to training on a huge graph with the knowledge that it is a union of several disconnected components.

\begin{figure}[htb]
	\centering
	\includegraphics[scale=0.4]{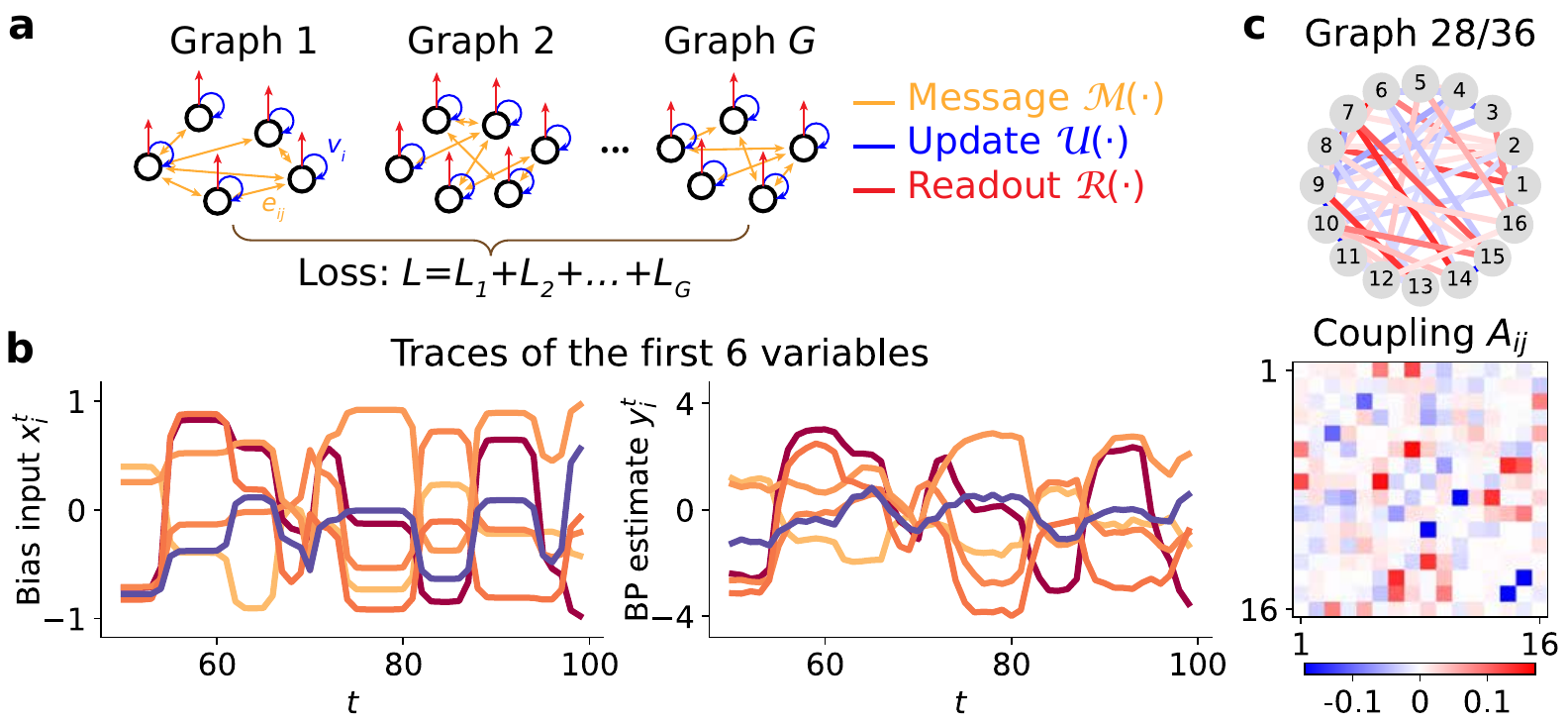}
	\caption{Schematic of multi-graph training and example traces of BP algorithm on a multivariate Gaussian distribution. (a) While each GNN has its own graph parameters, the canonical functions are shared by all. Canonical functions and graph parameters represent the dynamical and structural aspects of the system, respectively. (b) Bias sequence $b_i^t$ (\eq{eq:gauss.singleton}) serves as the input to GNN models, and the noisy BP estimation of marginal mean $\mu_i^t$ (\eq{eq:gauss.marginal}) is the target for GNN output to fit. Each variable is plotted with a different color. (c) One example of the 36 PGMs. Only the off-diagonal part of precision matrix $\vA$ is shown, characterizing the coupling among variables.}
	\label{fig:schematic.example}
\end{figure}

In this study, we prepare BP traces on multiple PGMs and simultaneously train multiple GNNs corresponding to each. The GNNs share the same canonical functions but different graph parameters for each individual PGM. The training objective function is simply the summation of loss function over all PGMs (\fig{fig:schematic.example}a). Each individual loss is designed to be proportional to the amount of data, therefore the trained canonical functions are naturally biased towards the PGMs with more training data.

We categorize the GNN parameters as dynamical parameters $\vTheta^\mathrm{D} = \vTheta_\mathcal{M} \cup \vTheta_\mathcal{U} \cup \vTheta_\mathcal{R}$ and structural parameters $\vTheta^\mathrm{S} = \{\ve_{ij}\} \cup \{\vv_i\}$. Denoting the number of GNN models as $G$, we need to learn one set of shared dynamical parameters $\vTheta^\mathrm{D}$, and $G$ sets of structure parameters $\vTheta^\mathrm{S}_1, \ldots, \vTheta^\mathrm{S}_G$ for each PGM respectively.

\subsection{Problem formulation}

We study the BP algorithm on a time-varying multivariate Gaussian distribution defined by
\begin{align}
	\phi_i^t(\theta_i) &= \exp\left(-\frac{a_i}{2}\theta_i^2+b_i^t \theta_i\right), \label{eq:gauss.singleton} \\
	\psi_{ij}(\theta_i, \theta_j) &= \exp\left(-J_{ij}\theta_i\theta_j\right). \label{eq:gauss.pairwise}
\end{align}

The joint distribution $p(\vtheta)$ can be rewritten as
\begin{equation}
	p(\vtheta) \propto \exp\left(-\frac{1}{2}\vtheta\trans\vA\vtheta + {\vb^t}\trans\vtheta\right), \label{eq:gauss.full}
\end{equation}
with $\vA_{ii} = a_i, \vA_{ij}=\vA_{ji}=J_{ij}$ as the precision matrix. The bias time series $\vb^t$ is a sequence of constant values with random duration following a Poisson process, while switching between periods are smoothed by a Hamming window.

Since the marginal distributions $p_i(\theta_i)$ are also Gaussian distributions, BP on Gaussian distribution \citep{Bickson2009} returns its single-variable means $\mu_i^t$ and standard deviations $\sigma_i^t$, \textit{i.e.},
\begin{equation}
	\hat{p}_i^t(\theta_i) \propto \exp\left(-\frac{1}{2{\sigma_i^t}^2}\left(\theta_i-\mu_i^t\right)^2\right). \label{eq:gauss.marginal}
\end{equation}
When converged, these means are exact but the standard deviations are approximate for loopy graphs \citep{weiss2001correctness}.
We use a damped version of the BP algorithm so that belief update is slower, and add processing noise at each time step (\cref{noisy.bp}). The damping parameter is adjusted so that transient dynamics makes up a significant portion of the full trial. When the precision matrix $\vA$ is constant, the estimated standard deviation $\sigma_i^t$ quickly converges to the true value $\frac{1}{\sqrt{a_i}}$ independent of any time-dependent bias terms, so we focus on $\mu_i^t$ only. An example trial of BP trace is shown in \fig{fig:schematic.example}b.

The inputs to GNN are $\vx_i^t = [b_i^t]$, and the targets for the GNN to match are defined as $\vy_i^t = [\mu_i^t]$. The GNN output $\vo_i^t$ (\eq{eq:canon.read}) is constructed to have the same dimension as $\vy_i^t$, in this case of dimension 1. We use the squared error $L = \sum_{i, t} \| \vo_i^t-\vy_i^t \|^2$ as the loss function. As introduced in \cref{sec:multi.train}, the losses on all PGMs are summed up as the overall objective function. In practice, each training batch is fetched from a randomly selected PGM. Our loss includes $L_2$ regularization on the structural parameters $\vTheta^\mathrm{S} = \{\ve_{ij}\} \cup \{\vv_i\}$, which turn out to be critical in disentangling $\vTheta^\mathrm{S}$ from $\vTheta^\mathrm{D}$ (see \cref{sec:graph.trans}).

BP traces for 36 random PGMs are generated, using different number of variables, number of trials and time duration for each. The graph density and coupling strengths are approximately the same for all PGMs. More details can be found in \cref{sec:trace.data}.

\section{Results}

\subsection{Architecture comparison}

The GNN architecture is determined by various hyper-parameters, including the underlying graph connectivity $\mathcal{E}$, the dimensions $D_v$, $D_e$, $D_s$, $D_m$, and hidden layer sizes of canonical functions $\mathcal{M}(\cdot)$ and $\mathcal{U}(\cdot)$. We study the effect of hyper-parameters by performing an extensive search of model configurations and comparing the fitting quality of the trained GNNs. We first define a set of possible values for each hyper-parameter, randomly choose combinations for a GNN architecture, and find the best architecture conditioned on each value to examine the effect of this hyper-parameter. For example, if the candidate values for edge dimension $D_e$ are $\{0, 2, 4, 8\}$, the best architectures found by hyper-parameter search conditioned on each $D_e$ value are compared.

\begin{figure}[htb]
	\centering
	\includegraphics[scale=0.4]{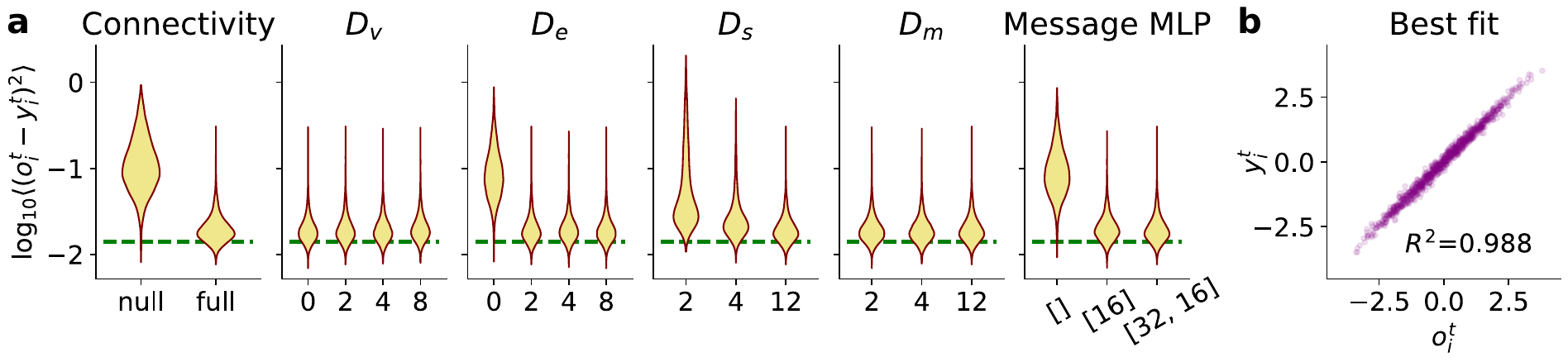}
	\caption{GNN architecture comparison and the best trained model. (a) The distribution of log mean squared errors (MSE) in the held-out testing set for the best GNN conditioned on each hyper-parameter value. Graph connectivity $\mathcal{E}$, vertex dimension $D_v$, edge dimension $D_e$, state dimension $D_s$, message dimension $D_m$ and the hidden layer size of message function $\mathcal{M}(\cdot)$ are examined. Dashed line is the baseline performance, which is the median of MSE if noiseless BP traces is treated as prediction. (b) BP target $\vy_i^t$ against GNN output $\vo_i^t$ of the best trained model.}
	\label{fig:arch.best.fit}
\end{figure}

The graph connectivity $\mathcal{E}$ is treated as a hyper-parameter because it is difficult to learn it in an end-to-end manner. Without any prior knowledge of the structure of a system, the only fair choices are a null graph ($\mathcal{E} = \varnothing$) or a complete graph ($\mathcal{E} = \{(i, j) \mid {(i, j)\in \mathcal{V}^2 \wedge i \neq j}\}$). The former (`null') assumes no coupling among variables, while the latter (`full') assumes all pairs of coupling are possible. Not surprisingly, `full' GNNs fit data better than `null' GNNs (\fig{fig:arch.best.fit}a), demonstrating the necessity of pairwise messages for explaining the traces in our BP example. It is worth mentioning that even the `null' GNNs produce good fit ($R^2=0.923$), because BP estimates $\mu_i^t$ in this example are largely determined by the singleton potentials $\phi_i(x_i)$ and only slightly `pulled' by other variables. The reason for using the moderately coupled system is to avoid numerical instability of BP.

We next examine the effects of model component dimensions, including $D_v$, $D_e$, $D_s$ and $D_m$. The results (\fig{fig:arch.best.fit}a) show that a non-zero edge dimension is critical in fitting the BP traces on Gaussian distribution; the real coupling is a scalar and thus of dimension 1. Surprisingly, the vertex dimension does not have to be greater than zero, even though the precision parameter $A_{ii}$ is a vertex attribute of dimension 1. We will discuss how to identify ground truth dimension through graph translators in \cref{sec:discuss}. There is a small benefit of increasing state dimension $D_s$, but no significant effect of message dimension $D_m$. It should be noticed that the best architecture for each conditioned value is usually different. For example, the best configuration is $(D_v=0, D_e=8)$ for $D_m=2$ and $(D_v=4, D_v=1)$ for $D_m=12$.

Lastly we examine the effect of message function complexity characterized by the hidden layer sizes in $\mathcal{M}(\cdot)$, and find that nonlinearity in the message is crucial for GNNs to fit well (\fig{fig:arch.best.fit}a). Due to the high cost of hyper-parameter search, we do not compare different update functions, but fix it as a GRU function modulated by the vertex parameter $\vv_i$. A more thorough search is left for future work.

To summarize, the results show strong dependency of fitting quality on graph connectivity, edge dimension and the message function nonlinearity, which indicates that in order to fit this dataset well, nonlinear messages between vertices are essential, and the GNN edges need to be parameterized.

\subsection{GNN training result}

One of the best architecture we find uses `full' connectivity, $D_e=2, D_v=2, D_h=12, D_m=12$ and a message function with one hidden layer of size 16. After a short initial burn-in period, the outputs of the trained GNNs faithfully reproduce the BP traces on the held-out testing set ($R^2=0.988$, \fig{fig:arch.best.fit}b). Not only do the GNN outputs reach the same equilibrium as BP targets within each input period, but the temporal profile at each input switch is also accurate. An example trial is shown in \cref{sec:best.fit} (\fig{fig:best.fit.example}).

\subsection{State and message manifold} \label{sec:manifold}

\begin{figure}[htb]
	\centering
	\includegraphics[scale=0.4]{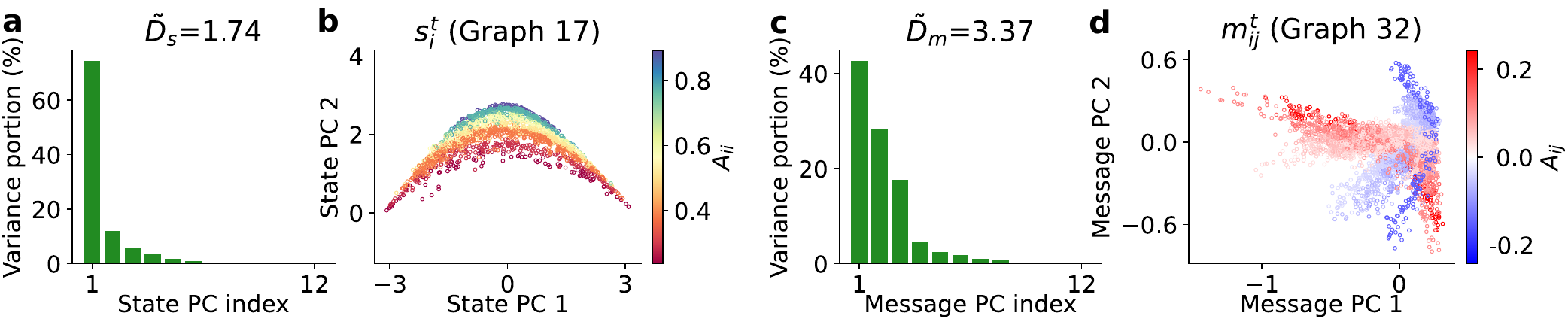}
	\caption{Manifold analysis of GNN states and messages. (a) PCA spectrum of states on all vertices in all GNNs. The effective dimension is $\tilde{D}_s\approx1.74$ (for $D_s=12$). (b) Projection of states $\vs_i^t$ of one GNN onto the space spanned by first two state PCs. Each point is colored by the parameter of vertex it belongs to, \textit{i.e.} the diagonal element of precision matrix $\vA$. (c) PCA spectrum of messages on all edges in all GNNs. The effective dimension is $\tilde{D}_m\approx3.37$ (for $D_m=12$). (d) Projection of messages $\vm_{ij}^t$ of one GNN onto the space spanned by first two message PCs. Each point is colored by the parameter of the edge it travels on, \text{i.e.} the coupling strength $A_{ij}$.}
	\label{fig:state.message.pca}
\end{figure}

We next analyze the states and messages of this well-trained GNN. We gathered these time series from all graphs and performed principal component analysis (PCA) on them. As expected, GNN states and messages only occupy a small portion of the high-dimensional space. The effective dimension of the manifold is defined as $\tilde{D} = \frac{\left(\sum_i{\lambda_i}\right)^2}{\sum_i{\lambda_i^2}}$,
where $\lambda_i$ denotes the variance of the $i$-th PC. When the state dimension is set to $D_s=12$, the effective dimension of state manifold is only $\tilde{D}_s \approx 1.74$ (\fig{fig:state.message.pca}a). We visualize GNN states in 2D and find they are organized by vertices. States at each vertex form its own curved 1D manifold, slightly separated for different vertices (\fig{fig:state.message.pca}b).

We perform the same analysis on pairwise messages as well. The effective dimension of messages is $\tilde{D}_m \approx 3.37$ for message dimension $D_m=12$ (\fig{fig:state.message.pca}c). Different manifolds occupy the message space with different orientations and offsets. Though messages gathered from all edges are not as structured as states, those on individual edges also trace out its own 1D manifold (\fig{fig:state.message.pca}d). We analyze the dimensionality of aggregated messages $\vm_i^t$ (\eq{eq:aggr.func}) and find they also lie approximately on a 1D manifold specific to each vertex (\cref{sec:agg.msg.pca}).

\subsection{Interpretable canonical functions}

\begin{figure}[htb]
	\centering
	\includegraphics[scale=0.4]{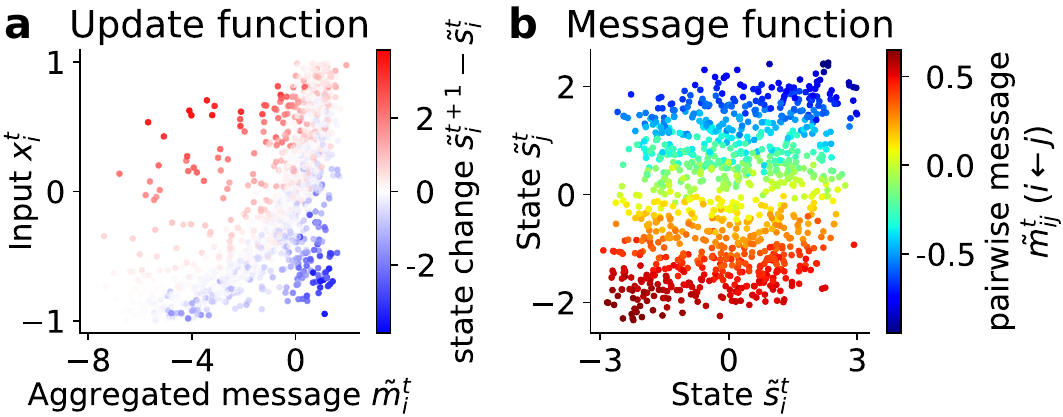}
	\caption{Visualizations of example update and message functions for a randomly selected vertex and edge. Since the high-dimensional states and messages lie on approximately 1-D manifolds, we plot these quantities according to their projection onto their first PCs. (a) State change as a function of aggregated message and the external input. (b) Pairwise message as a function of state at source vertex $i$ and target vertex $j$.}
	\label{fig:canonical.func}
\end{figure}

With a clearer picture of the GNN states and messages, we next analyze the learned canonical functions $\mathcal{U}(\cdot)$ and $\mathcal{M}(\cdot)$. We take advantage of the fact that the states and messages conditioned on a vertex or an edge approximately lie on a curved 1-D manifold (\cref{sec:manifold}). Therefore we can project the states or messages on the leading PC for that vertex or edge, and use this projection as a proxy to visualize how these functions depend on their inputs. Although inputs and outputs of $\mathcal{U}(\cdot)$ and $\mathcal{M}(\cdot)$ are vectors, we use scalar proxies as coordinates to plot heat maps of canoncial functions (\fig{fig:canonical.func}).

For a GNN vertex with parameter $\vv_i$, the update function $\mathcal{U}(\cdot)$ (\eq{eq:upd.func}) is a function of $\vs_i^t$, $\vx_i^t$ and $\vm_i^t$, in which the $\vm_i^t$ is the aggregated message for vertex $i$ at time $t$. The input $\vx_i^t$ is the external input, which is simply the local bias $b_i^t$ in this case. We denote the scalar proxies as $\tilde{s}_i^t$, $\tilde{s}_i^{t+1}$ and $\tilde{m}_i^t$ respectively. We then focus on the state change $\Delta\tilde{s}_i^t=\tilde{s}_i^{t+1}-\tilde{s}_i^t$ as a function of $\vx_i^t$ and $\tilde{m}_i^t$. \fig{fig:canonical.func}a shows one example, in which the state change $\Delta\tilde{s}_i^t$ is positive when $\tilde{m}_i^t$ is small and $\vx_i^t$ is large, and is negative otherwise. Hence $\Delta\tilde{s}_i^t$ encodes the discrepancy between aggregated message $\tilde{m}_i^t$ and the external input $\vx_i^t$. For a GNN edge with parameter $\ve_{ij}$, the message function $\mathcal{M}(\cdot)$ (\eq{eq:msg.func}) is a function of $\vs_i^t$ and $\vs_j^t$ that returns $\vm_{ij}^t$. Again we denote corresponding scalar proxies  as $\tilde{s}_i^t$, $\tilde{s}_j^t$ and $\tilde{m}_{ij}^t$. It appears that $\tilde{m}_{ij}^t$ changes monotonically with either $\vs_i^t$ or $\vs_j^t$, while depending mostly on the source vertex state $\tilde{s}_j^t$ (\fig{fig:canonical.func}b).

\subsection{Graph translator} \label{sec:graph.trans}

\begin{figure}[htb]
	\centering
	\includegraphics[scale=0.4]{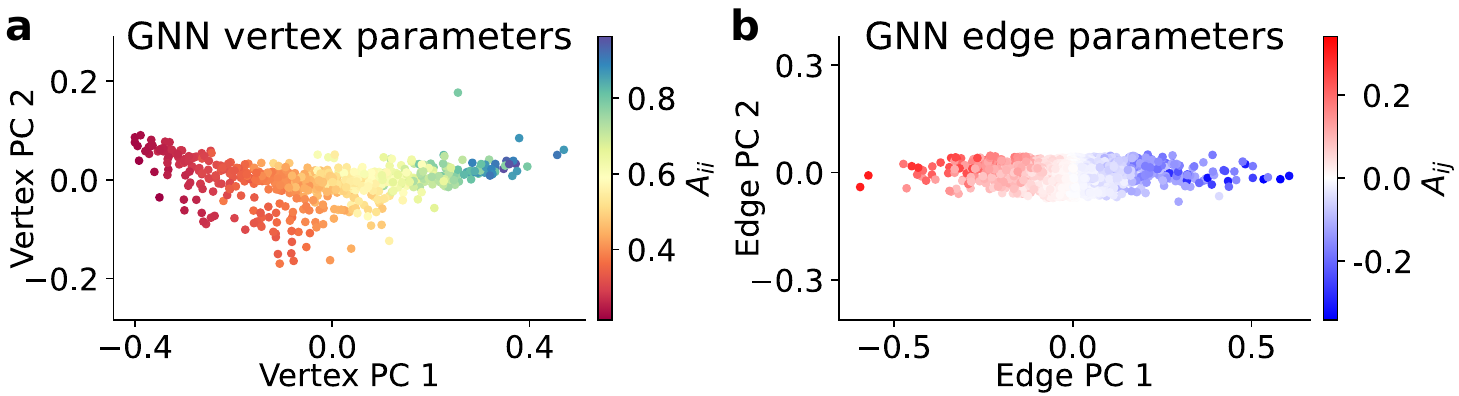}
	\caption{Low-dimensional structure of learned graph parameters. (a) Vertex parameters of dimension $D_v=2$ from all trained GNNs, colored according to the local precision parameter $A_{ii}$. (b) Edge parameters of dimension $D_e=2$ from all trained GNNs, colored according to the coupling strength parameter $A_{ij}$.}
	\label{fig:graph.param}
\end{figure}

The structural parameters $\vTheta^\mathrm{S} = \{\ve_{ij}\} \cup \{\vv_i\}$ of each individual GNN are learned independently, and they should relate to the true parameters of the corresponding PGMs. In a multivariate Gaussian distribution, a local precision parameter $A_{ii}$ is associated with each variable, and a coupling strength parameter $A_{ij}$ (\eq{eq:gauss.singleton}--\ref{eq:gauss.full}) is associated with each pair of variables. The hypothesis is that there exists a mapping between GNN vertex parameters $\vv_i$ and $A_{ii}$, as well as between GNN edge parameters $\ve_{ij}$ and $A_{ij}$. We will learn this mapping between GNN structural parameters and the static attributes of the target nonlinear dynamical system, and term it a `graph translator' because the conversion goes both ways.

Before learning the translators, we first look at the distribution of $\vv_i$ and $\ve_{ij}$. Though $D_v=D_e=2$ in the trained GNN, both vertex parameters and edge parameters approximately form a 1-D manifold. Moreover, locations on the two manifolds are continuously mapped to the corresponding attributes of PGM (\fig{fig:graph.param}).

The explicit structure that emerges in parameter space depends critically on the regularization of structural parameters during training. Such low-dimensional structure does not show up without the $L_2$ norm regularization on $\vv_i$ and $\ve_{ij}$. Well-trained but unregularized GNNs still give approximately the same good predictions of BP traces, but the effective dimensionality for vertex and edge parameters are very close to their embedding dimensions of $D_v$ and $D_e$ (\cref{sec:nowd.train}).

The graph translator we train is simply an MLP with two hidden layers, though it can be any regression model that predicts $A_{ii}$ from $\vv_i$ (vertex translator) or $A_{ij}$ from $\ve_{ij}$ (edge translator), or the opposite direction. To quantitatively evaluate graph translators, we divide all $G$ graphs into training, validation and testing sets. Only the original PGM attributes (\textit{e.g.} $A_{ii}$ and $A_{ij}$) of training and validation graphs will be used to learn the graph translator, and the testing graphs will be used to evaluate how well the translator behaves. The training graphs are used to train the translator directly, and the validation graphs are only used for early stopping. We also assume it is expensive to obtain the original graph attributes (\textit{e.g.} in neuroscience applications it is laborious to measure synapse strength by patch-clamping experiments), so we use only a subset of the data in training and validation graphs. Specifically, only 80\% of the vertices or edges are randomly selected for training the corresponding graph translator. All vertices and edges in the testing graphs are used for evaluation.

\begin{figure}[htb]
	\centering
	\includegraphics[scale=0.4]{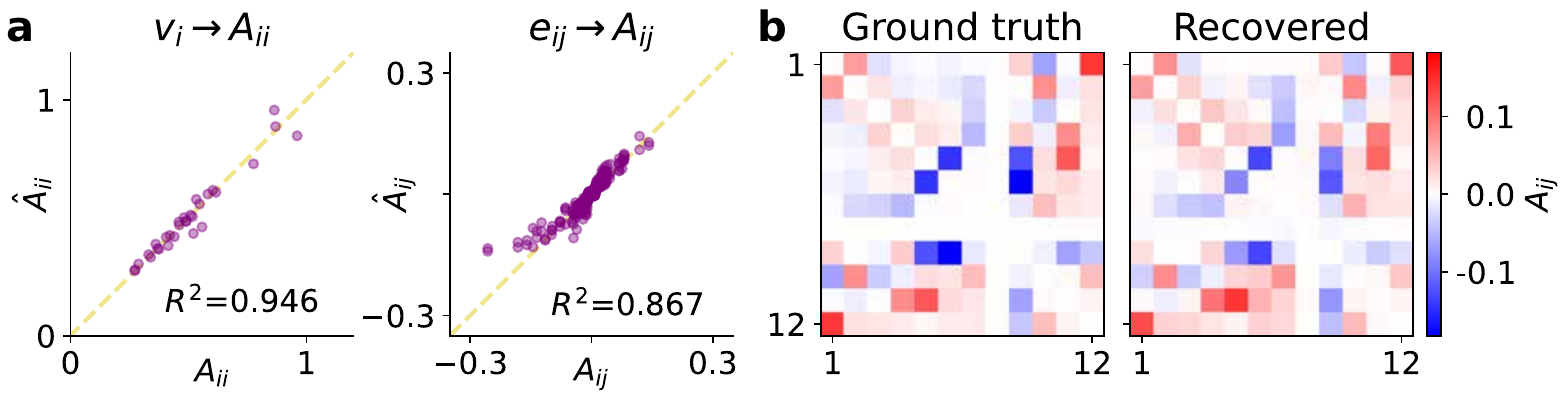}
	\caption{Translating from GNN graph parameters to precision matrix. (a) Both the vertex translator and edge translator predict attributes of PGM on the testing graphs. (b) The coupling matrix $A_{ij}$ ($i\neq j$) recovered by the edge translator closely resembles the ground truth. Diagonal part of $\vA$ is not shown for better visualization.}
	\label{fig:gnn2data}
\end{figure}

The translated local precision $A_{ii}$ and coupling strength $A_{ij}$ both match the ground truth well, with $R^2=0.946$ and $R^2=0.867$ respectively (\fig{fig:gnn2data}a). Recovered coupling matrix $A_{ij}$ looks similar to the ground truth (\fig{fig:gnn2data}b), revealing the correct interactions among different random variables. Since we do not enforce any symmetry about GNN edges $\ve_{ij}$, the recovered coupling matrix is not perfectly symmetric. It is straightforward to reveal the underlying structural connectivity between variables by thresholding the coupling strength (\cref{sec:connectivity.pred}).

Graph translators can also be used in the reverse direction to directly construct GNN models for a given PGM. The new GNN uses old dynamical parameter $\vTheta^\mathrm{D}$ with the new structural parameter $\vTheta^\mathrm{S}$ translated from precision matrix $\vA$. We compare the constructed GNNs with two control models. The first is the best trained colorless GNN with $D_v=D_e=0$, \textit{i.e.} vertices and edges are homogeneous in a graph. Colorless GNNs can be constructed for new PGM directly, without using a graph translator. The second control is the best-trained GNNs with the same parameter dimension as the constructed one, namely the GNNs we obtained earlier for the testing graphs.

\begin{figure}[htb]
	\centering
	\includegraphics[scale=0.4]{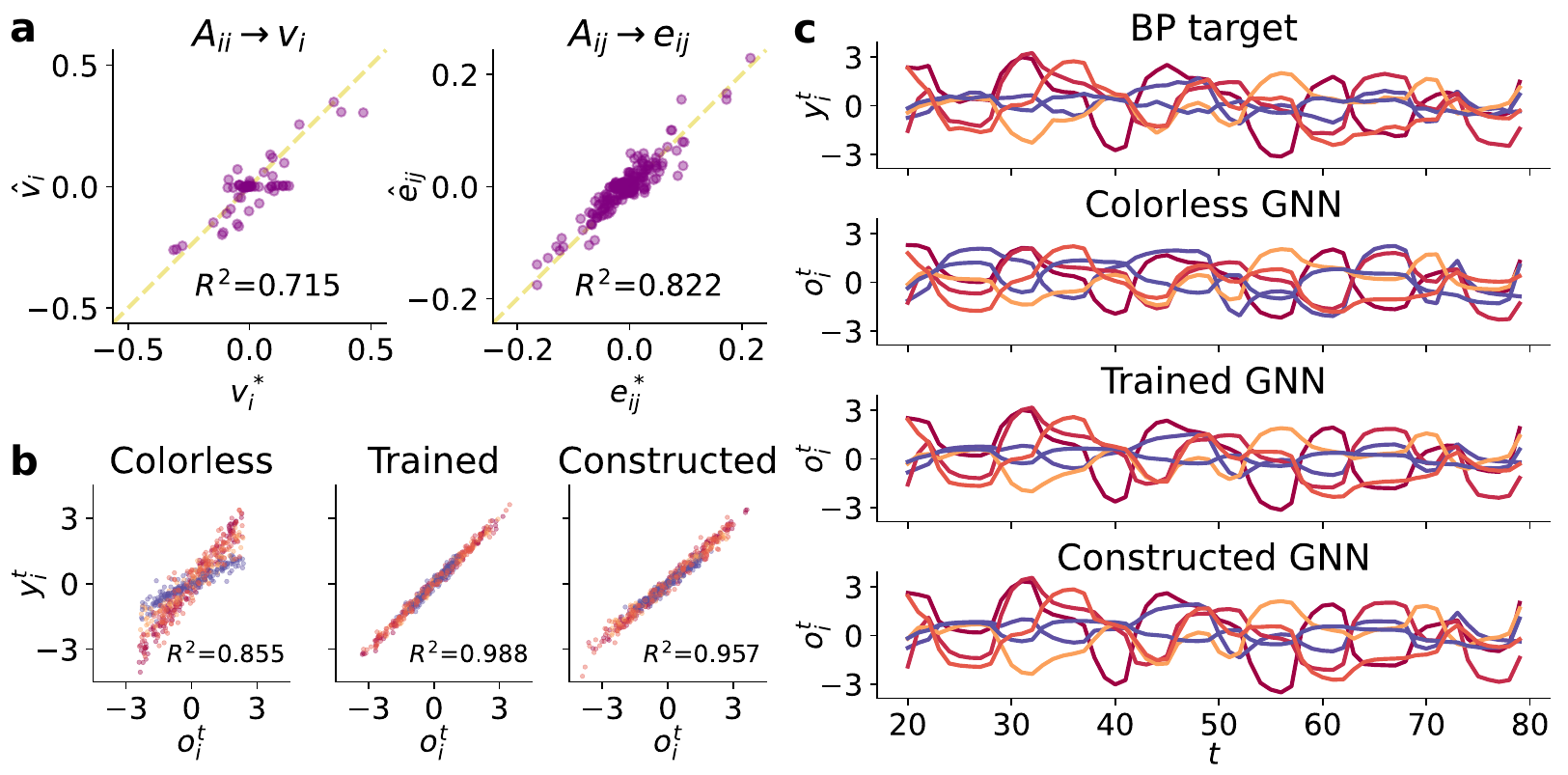}
	\caption{Translating a precision matrix to GNN graph parameters. (a) Comparison of graph parameters between the constructed GNN ($\hat{\vv}_i$, $\hat{\ve}_{ij}$) and the trained GNN ($\vv^*_i$, $\ve^*_{ij}$). (b) BP targets against GNN outputs for the colorless GNN ($D_v=D_e=0$), the trained GNN and the translated GNN. Data is colored by vertices. (c) An example trial comparing the BP target and GNN outputs, only 6 random variables are shown for clarity.}
	\label{fig:data2gnn}
\end{figure}

The graph parameters of constructed and trained GNNs are close to each other (\fig{fig:data2gnn}a). The constructed $\hat{\vv}_i$ and $\hat{\ve}_{ij}$ for the testing PGMs are close to the optimal values $\vv^*_i$ and $\ve^*_{ij}$ from the corresponding trained GNNs. Although the strong bias terms ($b_i^t$) ensure that even disconnected GNNs would capture some amount of the correct marginals, the colorless GNNs match true traces worst, indicating the benefits of parameterized edges. The trained GNNs match true traces best, since they have access to trace data of the new PGM. The translated GNNs, though never trained on these BP traces, can generate credible traces (\fig{fig:data2gnn}b). 

\section{Conclusion and discussions} \label{sec:discuss}

Using the belief propagation (BP) algorithm on Gaussian probabilistic graph models (PGM) as an example nonlinear dynamical system, we show that graph neural network (GNN) models can be trained on multiple instances of the same type of complex system. Via architecture search, we identify that messages generated by a nonlinear function of states and heterogeneous GNN edges are needed for fitting the BP traces well. We also show that the representation and canonical functions of GNN models have interpretations that are consistent with the principle of probabilistic inference. We further propose the novel concept of a graph translator that links graph parameters in GNN models to graph attributes of the original system. We show that the learned graph translator for Gaussian BP enables two types of generalization: recovering the precision matrix of a new Gaussian PGM from the traces of BP performing inference, and to construct GNN model for a new Gaussian PGM in order to reproduce BP algorithm without knowing its implementation. Unlike generalization to new inputs, such generalization to new instances of the same system builds on the successful disentanglement between dynamics and structure.

We find regularization is crucial to disentangle dynamical parameters $\vTheta^\mathrm{D} = \vTheta_\mathcal{M} \cup \vTheta_\mathcal{U} \cup \vTheta_\mathcal{R}$ and structural parameters $\vTheta^\mathrm{S} = \{\ve_{ij}\} \cup \{\vv_i\}$. When no regularization is used during GNN training, the learned $\vTheta^\mathrm{S}$ does not show clear structure despite the fact that multiple PGMs are trained together. $L_2$ norm regularization on vertex and edge parameters $\vv_i$ and $\ve_{ij}$ is used in our study, but other options might also work, \textit{e.g.} regularization on states $\vs_i$ or messages $\vm_{ij}$. We also observed that low-dimensional structure of GNN states $\vs_i$ and messages $\vm_{ij}$ is still present when no regularization is used (\cref{sec:nowd.train}).

Though BP on Gaussian PGM is an important probabilistic inference algorithm, its dynamics are undoubtedly still simple. In future work we would like to test our framework in other complex systems such as celestial mechanics, epidemiology, population ecology, and neuroscience. More advanced GNN architecture is perhaps necessary for these applications, such as stacking GNN layers or attention-based message aggregation. One straightforward next step is to study PGMs with more complicated couplings, for instance a multivariate von Mises distribution whose pairwise potential $\psi_{ij}(\theta_i, \theta_j)$ is specified by 4 free parameters. We predict that GNN edge dimension $D_e$ has to be at least 4 to faithfully model BP traces on such PGMs.

In this study, we make zero assumption about the connectivity of the underlying graph, which leaves us to choose either a null graph or a fully connected one. However the connectivity can also be learned in principle if proper inductive bias is imposed, \textit{e.g.} a sparsity prior, proximity rules, etc. In fact, one can try to estimate the connectivity from the learned fully connected edge parameters $\ve_{ij}$ and use these estimates as the graph's edges $\mathcal{E}$ for the next iteration of GNN training. Approaches that learn connectivity end-to-end should also be explored.

From fitting quality alone, we obtain a basic picture on the dimensions of the complex system (\fig{fig:arch.best.fit}), but identification of the exact values of $D_v$, $D_e$, $D_s$ and $D_m$ is still an open question. Preliminary data suggests that it is possible to identify them using generalization performance on new instances (\fig{fig:data2gnn}) as a metric. Only the GNNs with correct assumptions about dimensions can accurately predict system behavior on a new instance, because either underestimation or overestimation will result `incorrectly' constructed GNNs. We will examine the results further, also include $D_v=1$ and $D_e=1$ in the architecture search.


\section*{Acknowledgements}
The authors thank Rajkumar Raju, Yicheng Fei, and KiJung Yoon for helpful conversations. This work was supported in part by NSF CAREER grant 1552868 to XP, NSF NeuroNex grant 1707400 to AT and XP, an award from the McNair Foundation to XP, AFOSR grant FA9550-21-1-0422 in the Cognitive and Computational Neuroscience program to XP, and the Intelligence Advanced Research Projects Activity (IARPA) via Department of Interior/Interior Business Center (DoI/IBC) contract number D16PC00003 to AT and XP. The U.S. Government is authorized to reproduce and distribute reprints for Governmental purposes notwithstanding any copyright annotation thereon. Disclaimer: the views and conclusions contained herein are those of the authors and should not be interpreted as necessarily representing the official policies or endorsements, either expressed or implied, of IARPA, DoI/IBC, or the U.S. Government.

Disclosed interests: AT and XP are co-founders of Upload AI, LLC.

\appendix

\section{Meta-MLP} \label{sec:meta-mlp}

We design a family of MLPs organized by meta-parameters, called meta-MLP or mMLP for short. In an original MLP, the input vector $\vx$ is defined as layer 0 activation $\vx^0$. $L$ sequential perceptrons transform activation from one layer to the next by
\begin{equation}
	\vx^{l} = f^l \left( \vW^l \vx^{l-1} + \vb^l \right),
\end{equation}
in which $\vW^l$ and $\vb^l$ are the weight and bias of layer $l$ respectively, and $f^l(\cdot)$ is the nonlinear activation function of layer $l$. The $L$-th layer activation $\vx^L$ is defined as the output of the MLP,
\begin{equation}
	\mathrm{MLP}(\vx; \vTheta) \equiv \vx^L,
\end{equation}
with $\vTheta = \{ \vW^1, \vb^1, \ldots, \vW^L, \vb^L \}$.

In an mMLP with meta-parameter $\vzeta$, each layer is defined as
\begin{equation}
	\vx^{l+1} = f^l \left( \vW^l \left[ \begin{array}{c} \vx^l \\ \vzeta \end{array} \right] + \vb^l \right). \label{eq:mMLP.layer}
\end{equation}
In \eq{eq:mMLP.layer}, the meta-parameter $\vzeta$ modulates each layer transformation by acting as an extra input. The output of the mMLP is defined as the last layer activation
\begin{equation}
	\mathrm{mMLP}(\vx, \vzeta; \vTheta) \equiv \vx^L.
\end{equation}

Optionally, batch normalization layers can be added right before each nonlinear activation function. When a batch normalization layer with trainable affine transformation is added, the bias $\vb^l$ will be removed.

In this study, all layers in an mMLP except the last one use the nonlinear ELU activation, \textit{i.e.}
\begin{equation}
	f^l(z) = \left\{ \begin{array}{ll} z & z\geq 0, \\ e^z-1 & z<0, \end{array} \right.
\end{equation}
for $1\leq l < L$. The activation function of last layer is the identity function $f^L(z) = z$.

\section{Canonical functions} \label{sec:canon.func}

\subsection{Message function} \label{sec:msg.func}

In this study, the message function $\mathcal{M}(\cdot)$ is defined as
\begin{equation}
	\mathcal{M}(\vs_{i}^t, \vs_{j}^t; \ve_{ij}, \vTheta_\mathcal{M}) \equiv \mathrm{mMLP} \left( \left[ \begin{array}{c} \vs_i^t \\ \vs_j^t \end{array} \right], \ve_{ij} ; \vTheta_\mathcal{M}
	\right).
\end{equation}
While message parameter $\vTheta_\mathcal{M}$ is shared across all edges, meta-parameter $\ve_{ij}$ are different for different edges.

\subsection{Update function} \label{sec:upd.func}

We use a modified version of gated recurrent unit (GRU) function \citep{Cho2014} as the update function $\mathcal{U}(\cdot)$ (\eq{eq:msg.func}). The form of original GRU update is
\begin{align*}
	\vz^t &= \sigma_g \left(\vW_z \vx^t + \vU_z \vs^t + \vb_z\right), \\
	\vr^t &= \sigma_g \left(\vW_r \vx^t + \vU_r \vs^t + \vb_r\right), \\
	\vs^{t+1} &= (1-\vz^t) \circ \vs^t + \vz^t \circ \sigma_s \left(\vW_s \vx^t + \vU_s \left( \vr^t \circ \vs^t \right) + \vb_h\right),
\end{align*}
in which $\vx^t$ and $\vs^t$ are the external input and hidden state at time $t$ respectively. $\vz^t$ and $\vr^t$ are the update gate and reset gate vectors, while $\sigma_g(\cdot)$ and $\sigma_s(\cdot)$ are nonlinear functions typically chosen as logistic function and hyperbolic tangent function respectively.

While original GRU uses perceptrons for gates and state updates, we replace them by meta-MLPs with vertex parameter $\vv_i$. We do not append nonlinear activation to the last layer of mMLP (\cref{sec:meta-mlp}), hence the gating nonlinearities $\sigma_g$ and $\sigma_s$ are kept. Another difference is each vertex receives not only external input, but also aggregated messages from neighbors. The update function $\mathcal{U}(\cdot)$ (\eq{eq:upd.func}) is defined through
\begin{align}
	\vz_i^t &= \sigma_g \left(\mathrm{mMLP}\left(\left[ \begin{array}{c} \vi_i^t \\ \vm_i^t \\ \vs_i^t \end{array} \right], \vv_i; \vTheta_{\mathcal{U}^z}\right)\right), \label{eq:mmlp.zgate} \\
	\vr_i^t &= \sigma_g \left(\mathrm{mMLP}\left(\left[ \begin{array}{c} \vi_i^t \\ \vm_i^t \\ \vs_i^t \end{array} \right], \vv_i; \vTheta_{\mathcal{U}^r}\right)\right), \label{eq:mmlp.rgate} \\
	\vs_i^{t+1} &= (1-\vz_i^t) \circ \vs_i^t + \vz_i^t \circ \sigma_s \left(\mathrm{mMLP}\left(\left[ \begin{array}{c} \vi_i^t \\ \vm_i^t \\ \vs_i^t \end{array} \right], \vv_i; \vTheta_{\mathcal{U}^s}\right)\right). \label{eq:mmlp.gru.state}
\end{align}
The update parameter is defined as $\vTheta_\mathcal{U} = \vTheta_{\mathcal{U}^z} \cup \vTheta_{\mathcal{U}^r} \cup \vTheta_{\mathcal{U}^s}$.

\subsection{Readout function}

We choose a simple linear function as canonical readout $\mathcal{R}(\cdot)$ (\eq{eq:canon.read}),
\begin{equation}
	\mathcal{R}(\vs_i^t;\vTheta_\mathcal{R})=\vW_\mathcal{R} \vs_i^t + \vb_\mathcal{R},
\end{equation}
with readout parameter $\vTheta_\mathcal{R} = \{ \vW_\mathcal{R}, \vb_\mathcal{R} \}$.

\section{Noisy belief propagation algorithm} \label{sec:noisy.bp}

We use belief propagation (BP) algorithm to estimate the marginal distribution of all variables. We define the message from vertex $j$ to $i$ as $m_{ij}(\theta_i)$, namely the belief about $i$ from $j$. At each iteration, BP updates $m_{ij}(\theta_i)$ to
\begin{gather}
	m_{ij}(\theta_i) = \frac{\tilde{m}_{ij}(\theta_i)}{\sum_{\theta_i}\tilde{m}_{ij}(\theta_i)}, \\
	\tilde{m}_{ij}(\theta_i) = \sum_{\theta_j} \left( \phi_j(\theta_j) \psi_{ij}(\theta_i, \theta_j) \prod_{k \in \mathrm{N}(j) \backslash i} m_{jk}(\theta_j) \right),
\end{gather}
in which $\mathrm{N}(j)$ is the set of neighbors of $j$. The marginal distribution of $\theta_i$ is
\begin{gather}
	p_i(\theta_i) = \frac{\tilde{p}_i(\theta_i)}{\sum_{\theta_i}\tilde{p}_i(\theta_i)}, \\
	\tilde{p}_i(\theta_i) = \phi_i(\theta_i) \prod_{j\in \mathrm{N}(i)}m_{ij}(\theta_i).
\end{gather}
We explicitly normalize messages and estimated marginal distributions at each iteration.

BP is not guaranteed to converge on loopy graphs, even when it converges it may not converge to the true marginal distribution. However if damping update is used, BP usually gives a stable good approximation and can be considered as a valid inference algorithm. We denote the damping coefficient as $\gamma$. BP updates messages in logarithm domain following
\begin{gather}
	m^t_{ij}(\theta_i) = \frac{\tilde{m}^t_{ij}(\theta_i)}{\sum_{\theta_i}\tilde{m}^t_{ij}(\theta_i)}, \\
	\ln \left(\tilde{m}_{ij}^{t+1}(\theta_i)\right) = \gamma \ln \left(m_{ij}^t(\theta_i)\right) + (1-\gamma) \ln \left(\sum_{\theta_j} \left( \phi_j^t(\theta_j) \psi_{ij}(\theta_i, \theta_j) \prod_{k \in \mathrm{N}(j) \backslash i} m_{jk}^t(\theta_j) \right)\right), \label{eq:damping_update}
\end{gather}
with superscript $t$ marks the time step in BP. Here the singleton potential $\phi_i^t(\theta_i)$ changes over time. Similarly, the estimated marginal distribution is also dynamic,
\begin{gather}
	p^t_i(\theta_i) = \frac{\tilde{p}^t_i(\theta_i)}{\sum_{\theta_i}\tilde{p}^t_i(\theta_i)}, \label{eq:bp_output} \\
	\tilde{p}_i^t(\theta_i) = \frac{1}{Z} \phi_i^t(\theta_i) \prod_{j \in \mathrm{N}(i)} m_{ij}^t (\theta_i).
\end{gather}

We additionally add processing noise to the inference algorithm to mimic a physical system. Since the message by definition is non-negative, we use additive noise in logarithmic domain to distort each update step. The complete form is
\begin{equation}
	\ln \left(m_{ij}^t(\theta_i)\right) + (1-\gamma) \ln \left(\sum_{\theta_j} \left( \phi_j^t(\theta_j) \psi_{ij}(\theta_i, \theta_j) \prod_{k \in \mathrm{N}(j) \backslash i} m_{jk}^t(\theta_j) \right)\right)+n_i^t, \label{eq:noisy_update}
\end{equation}
in which $n_i^t \sim \mathcal{N}(0, \sigma_\mathrm{n}^2)$ is independent Gaussian noise with variance $\sigma_\mathrm{n}^2$.

\section{Trace data} \label{sec:trace.data}

BP traces are generated for 36 PGMs. For each PGM, its graph size is randomly sampled from $\{12, 14, 16, 18\}$, the BP duration is randomly sampled from $\{80, 100, 120\}$, and the number of trials is randomly sampled from $\{1000, 1250, 1500\}$. Precision matrix of each PGM is a random positive-definite matrix, generated by applying random rotations on a subset of indices starting from a diagonal matrix until desired density is reached (similar to `sprandsym' function in MATLAB). The reciprocal condition number is 0.2, and the desired density is 60\% (estimated using a threshold of $\epsilon=0.01$ when the starting diagonal matrix has maximum value of 1).

The full dataset includes approximately 75 million data points (graph size $\times$ duration $\times$ number of trials, summed up for all 36 PGMs). 90\%, 5\% and 5\% of the data are used for training, validation and testing respectively.

\section{Example trial of the best GNN} \label{sec:best.fit}

\begin{figure}[htb]
	\centering
	\includegraphics[scale=0.45]{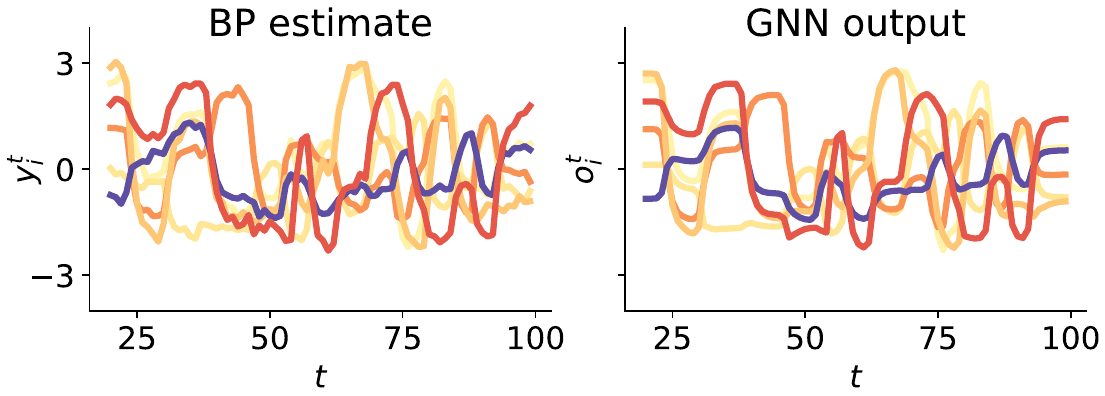}
	\caption{An example trial of the first six random variables of one PGM. The input sequence is from a held-out testing set.}
	\label{fig:best.fit.example}
\end{figure}

An example trial of the best fit GNN (\fig{fig:arch.best.fit}b) is shown in \fig{fig:best.fit.example}.

Since no processing noise is introduced in GNN, its output is smoother than the target traces given by noisy BP algorithm. The trained GNN captures main component of observed dynamical data, effectively removing the noise within.

\section{PCA on aggregated messages} \label{sec:agg.msg.pca}

PCA is performed on the aggregated messages $\vm_i^t$ (\eq{eq:aggr.func}) for the trained GNN. Similar to the geometry of states $\vs_i^t$ (\fig{fig:state.message.pca}), aggregated messages for each vertex also lie on a curved 1-D manifold in the high-dimensional space.

\begin{figure}[htb]
	\centering
	\includegraphics[scale=0.45]{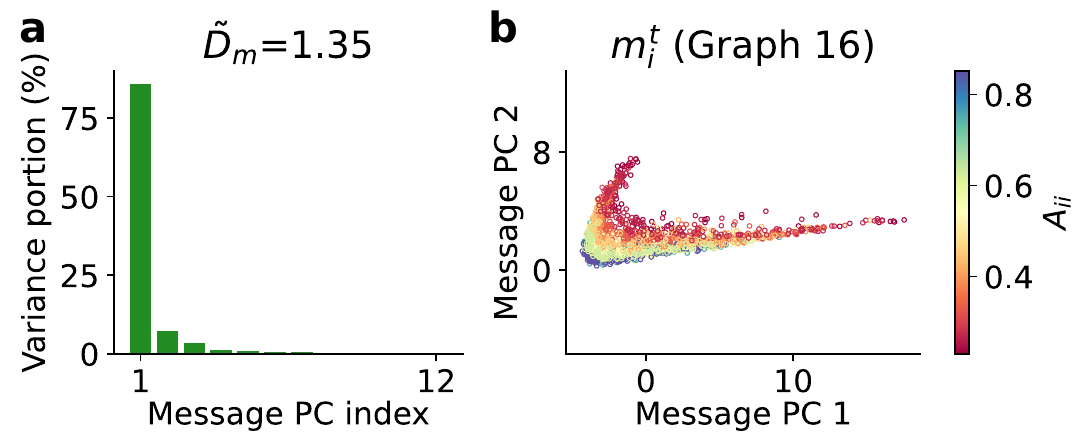}
	\caption{Manifold analysis of aggregated messages. (a) PCA spectrum of aggregated messages $\vm_i^t$ in the example GNN. (b) 2D visualization of $\vm_i^t$, colored by the precision matrix of each vertex.}
	\label{fig:agg.msg.pca}
\end{figure}

\section{Training with no regularization} \label{sec:nowd.train}

When no regularization is added on structural parameters $\vv_i$ and $\ve_{ij}$, the trained GNN can still predict BP traces well. One of the best architectures we found is $D_e=8, D_v=4, D_h=12, D_h=12$ and a message function with two hidden layers of size [32, 16]. Though the its fitting performance is high ($R^2=0.989$), PCA on the learned $\vv_i$ and $\ve_{ij}$ does not reveal low-dimensional manifold.

\begin{figure}[htb]
	\centering
	\includegraphics[scale=0.45]{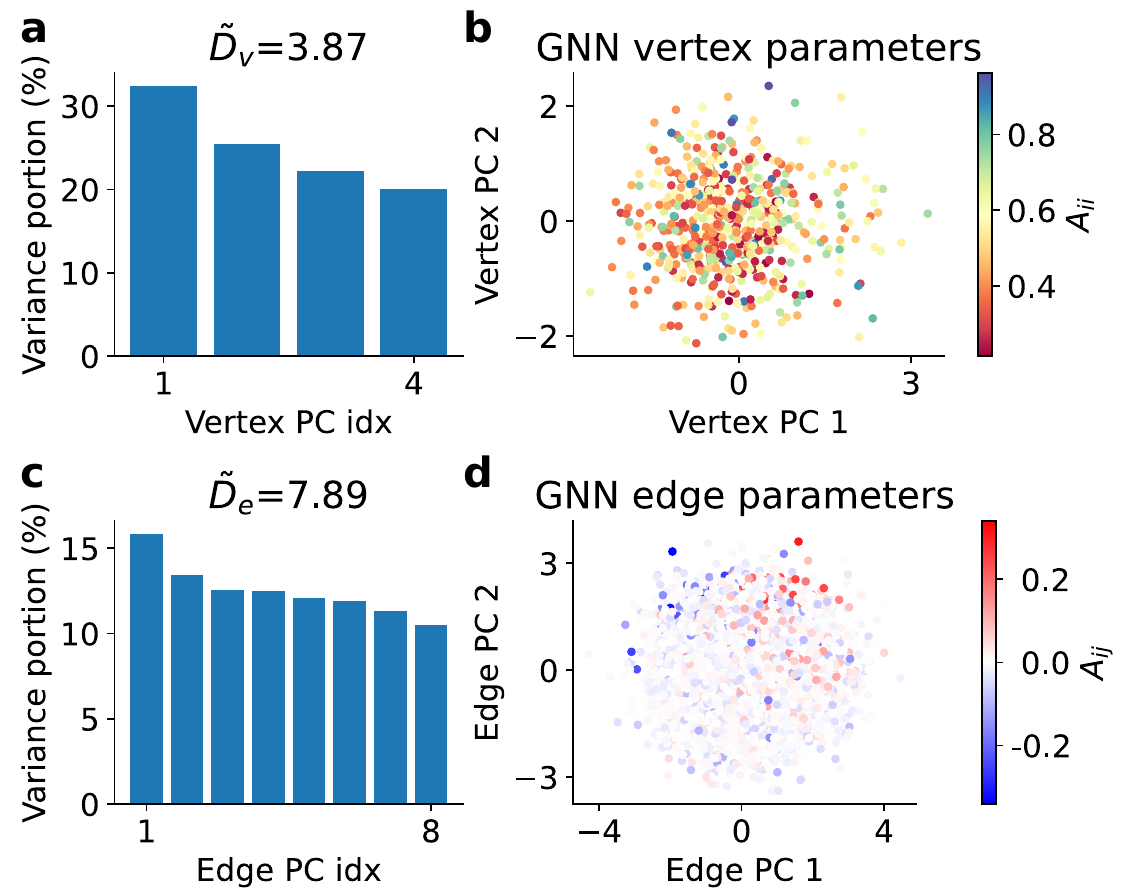}
	\caption{Learned structural parameters with no regularization. (a) PCA spectrum of $\vv_i$ when $D_v=4$. (b) 2D visualization of $\vv_i$ colored by $A_{ii}$. (c) PCA spectrum of $\ve_{ij}$ when $D_e=12$. (d) 2D visualization of $\ve_{ij}$ colored by $A_{ij}$.}
	\label{fig:graph.param.nowd}
\end{figure}

However, PCA on the states and pairwise messages of the trained GNNs still show low-dimensional structure, similar to \fig{fig:state.message.pca}. In fact, the messages seem to be more organized as the 1-D manifolds belonging to different edges are aligned parallel.

\begin{figure}[htb]
	\centering
	\includegraphics[scale=0.45]{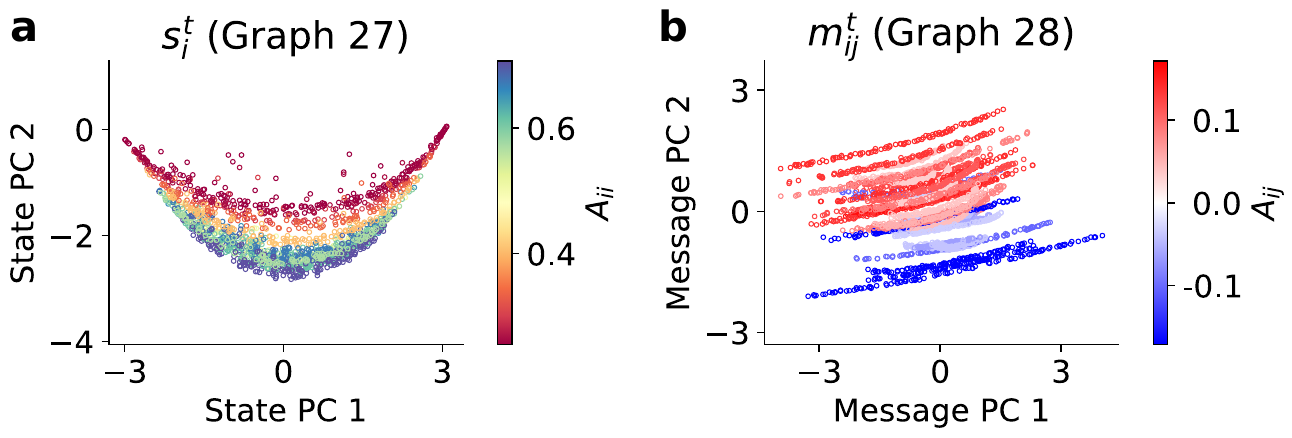}
	\caption{PCA of GNN states and messages when no regularization is used during training. (a) Projection of states $\vs_i^t$ onto the space spanned by its first two PCs, colored by precision parameter $A_{ii}$ (b) Projection of messages $\vm_{ij}^t$ onto the space spanned by its first two PCs, colored by coupling strength $A_{ij}$.}
	\label{fig:state.message.nowd}
\end{figure}

\section{Connectivity prediction} \label{sec:connectivity.pred}
\begin{figure}[htb]
 	\centering
 	\includegraphics[scale=0.4]{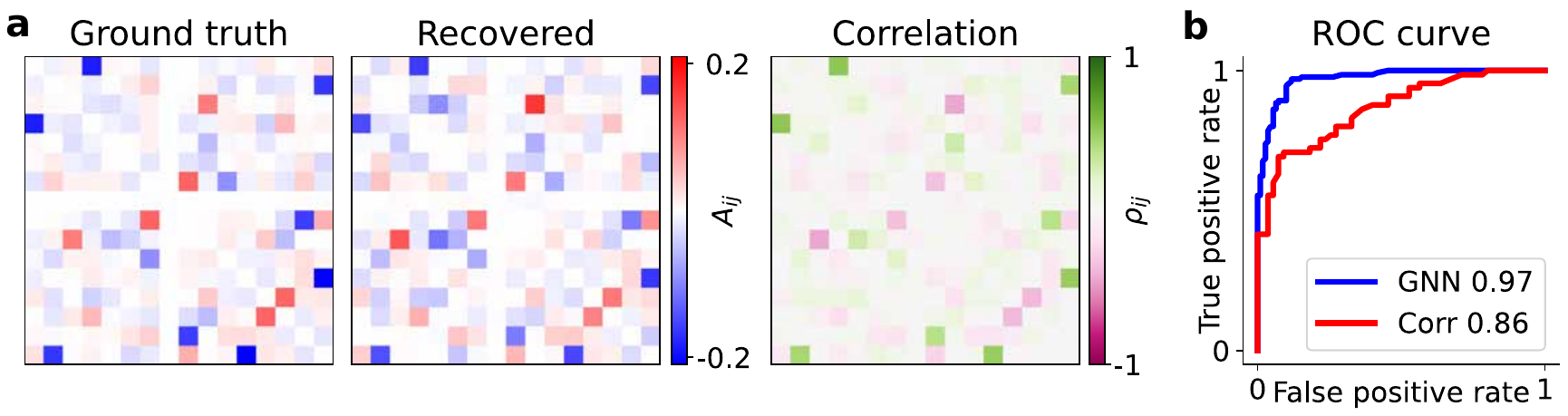}
 	\caption{Predicting underlying graph. Prediction based on correlation is not as accurate as from GNN recovery.}
 	\label{fig:roc}
 \end{figure}

 When two nodes are correlated, it does not necessarily indicate there is an edge connecting them. The underlying graph we use is not fully connected, \textit{i.e.} some values of the coupling matrix $A_{ij}$ is set to 0. We predict the connectivity by thresholding either the activity correlation matrix $\rho_{ij}$ or the coupling matrix $\hat{A}_{ij}$ recovered by edge translator (\fig{fig:roc}a), and plot the ROC curves for both case (\fig{fig:roc}b). Area under the curve (AUC) is larger for the prediction from recovered coupling matrix.

\end{document}